\documentclass{article}

 \usepackage[preprint]{sty_mod}

\author{%
  Vatsal Baherwani\\
  New York University\\
  \texttt{vatsalbaherwani@nyu.edu} \\
  \And
  Tom Goldstein \\
  University of Maryland \\
  \texttt{tomg@umd.edu} \\
  \And
  Ashwinee Panda \\
  TogetherAI \\
  \texttt{apanda@together.ai} \\
}

\usepackage[utf8]{inputenc} %
\usepackage[T1]{fontenc}    %
\usepackage{hyperref}       %
\usepackage{url}            %
\usepackage{booktabs}       %
\usepackage{amsfonts}       %
\usepackage{nicefrac}       %
\usepackage{microtype}      %
\usepackage{xcolor}         %

\usepackage{graphicx}
\usepackage{amsmath}
\usepackage[table]{xcolor}
\usepackage{bbm}
\usepackage{multirow}

\title{Not All LLM Reasoning is Visible in the Chain-of-Thought}

\begin{document}

\maketitle

\begin{abstract}
A key question for AI safety is whether a language model expresses all of its reasoning in its output tokens. We demonstrate a concrete failure mode where frontier models exhibit \textit{invisible reasoning} by leveraging semantically irrelevant filler tokens to improve performance on synthetic reasoning tasks. We evaluate 13 frontier language models across three tasks and find that many models benefit significantly from filler tokens, with accuracy improvements of up to 13 percentage points. The benefit depends on which tokens are used and differs across models.
We further show that filler tokens enable Claude Opus~4.5 to satisfy a hidden modular arithmetic constraint without sacrificing accuracy on its primary task, demonstrating that invisible reasoning can serve objectives entirely invisible to CoT monitoring. Reinforcement learning gives Qwen3-235B strong preferences over filler token content, but neither RL nor supervised fine-tuning produces a filler token benefit that persists at test time. Our results indicate that frontier models already perform consequential computation with no interpretable trace in their output tokens.
\end{abstract}

\section{Introduction}
Chain-of-thought (CoT) monitoring offers a tractable approach to auditing increasingly capable language models~\citep{korbak2025chainthoughtmonitorabilitynew}. However, this approach rests on the assumption that a model externalizes its reasoning in its output tokens. 
Prior work shows that a CoT may not be a faithful representation of a model's internal reasoning process~\citep{turpin2023languagemodelsdontsay, lanham2023measuringfaithfulnesschainofthoughtreasoning, arcuschin2025cotnotfaithful, chen2025reasoningmodelsdontsay, boppana2026reasoningtheaterdisentanglingmodel}. %
In this paper, we take this direction to the extreme and ask {\em whether reasoning is consistently observable in latent space, even in the absence of any meaningful tokens at all}. A forward pass through a model comprises billions of computations and thousands of latent vector representations before collapsing this information into a single output token. Given the richness of latent representations and the comparatively low information content of a single token, we expect internal reasoning processes to emerge that do not require a CoT.

This paper studies the existence of invisible reasoning from a fundamental perspective. However, developers also have practical incentives to elicit invisible reasoning in their models. Non-CoT performance is a standard benchmark criterion, and serving costs favor concise outputs. A model that leverages prefilled computation to improve accuracy can serve responses more efficiently. Recently, \citet{ramji2026thinkingwordsefficientlatent} propose training a model with an ``abstract'' CoT for more efficient inference. Invisible reasoning is also resistant to distillation, since meaningful computation does not appear in the output trace. We therefore expect future models to be explicitly optimized for latent computation, making our characterization of when and how it emerges directly relevant to practice. As we show in this paper, \textbf{frontier models already display the capability for invisible reasoning} without any additional explicit optimization.

We demonstrate that frontier models leverage semantically irrelevant \textit{filler tokens}: fixed token sequences that allow the model to do more computation before answering. These sequences are the same for every question and answer, so the reasoning they induce cannot be distinguished in token space. %
We consider this an example of \emph{invisible reasoning}, where a model performs computation that is invisible to an observer reading its outputs. We formalize invisible reasoning through three diagnostic criteria and show that the capability already exists in recent frontier models (Section~\ref{sec:filler}, \ref{sec:frontier}).
We then show that providing filler tokens, and therefore more space for invisible reasoning, enables Claude Opus~4.5 to fulfill a simple hidden goal alongside a primary task 
(Section~\ref{sec:frontier}). These experiments demonstrate the capability in synthetic settings; we do not claim that filler tokens evade a monitor in a realistic misbehavior setting.
We also study whether invisible reasoning can be strengthened through explicit training (Section~\ref{sec:rl}). Through RL, Qwen3-235B develops strong preferences over filler token content, but the filler token benefit does not persist at test time, and supervised fine-tuning fails to transfer it. Our results show that \textbf{models can perform consequential computation with no interpretable trace in their output tokens}.

\section{Related Work}

\paragraph{Chain-of-Thought Reasoning}
Chain-of-thought prompting~\citep{wei2023chainofthought, kojima2023largelanguagemodelszeroshot} is a standard technique for eliciting complex reasoning in transformer-based language models. CoT expands the computational capabilities of transformers \citep{merrill2024expressivepowertransformerschain}, particularly by improving their ability to solve serial problems \citep{li2024chainthoughtempowerstransformers}. 

\paragraph{CoT Faithfulness}
While chain-of-thought reasoning improves the problem solving capabilities of language models, it is unclear whether CoT accurately reflects a model's internal computation. LLMs are known to misrepresent their reasoning in CoT, often providing a post-hoc explanation of a predetermined answer rather than a genuine description of their internal reasoning~\citep{turpin2023languagemodelsdontsay, lanham2023measuringfaithfulnesschainofthoughtreasoning, arcuschin2025cotnotfaithful, chen2025reasoningmodelsdontsay}. \citet{korbak2025chainthoughtmonitorabilitynew} argue that although CoT may be unfaithful in simple settings, solving complex problems often requires LLMs to externalize their reasoning. However, recent work shows that language models can learn to reason in abstract, uninterpretable languages through reinforcement learning~\citep{ramji2026thinkingwordsefficientlatent}. Moreover, \citet{boppana2026reasoningtheaterdisentanglingmodel} provide concrete evidence of performative reasoning in CoT, showing that a model's final answer can be decoded from activations far earlier in the CoT than a monitor can detect it. These findings indicate that LLMs do not necessarily need to output interpretable explanations of their reasoning to leverage the computational benefits of CoT.

\paragraph{Steganography in Language Models.}
A key concern for AI oversight is whether models can hide information in their outputs. \citet{roger2023preventinglanguagemodelshiding} show that language models can be trained to encode hidden reasoning steps in output text imperceptible to readers, and that this risk grows with model scale. \citet{mathew2024hiddeninplaintext} further demonstrate that steganographic collusion can emerge unintentionally from misspecified reward incentives, and that standard mitigations such as paraphrasing are insufficient to prevent it. Invisible reasoning is a related phenomenon where computation is hidden in latent representations rather than in the statistical properties of output text.

\paragraph{Implicit and Internalized Reasoning}
Unfaithful CoT raises the question of whether models can reason effectively without producing interpretable intermediate steps at all. \citet{deng2023implicitchainofthought} show that distillation from an explicit-CoT teacher can train models to reason across layers rather than tokens, and \citet{deng2024explicitcotimplicitcot} extend this by progressively removing CoT steps during fine-tuning, achieving over 50\% accuracy on GSM8K with no visible reasoning. \citet{hao2024coconut} extend this approach further by feeding the model's last hidden state directly back as the next input embedding, enabling reasoning in a continuous latent space uninterpretable to humans. \citet{su2025tokenassorted} experiment with mixing discrete latent tokens and text tokens to produce partially opaque reasoning traces. These approaches require explicit training to suppress or replace the CoT, whereas we find that invisible reasoning emerges in standard pretrained models.

\paragraph{Invisible Reasoning and Filler Tokens}
Even without training for implicit reasoning, extra tokens in the context can provide computational benefit. \citet{pfau2024letsthinkdotdot} show that semantically meaningless filler tokens improve performance on parallelizable tasks, but only when the model is specifically trained with filler tokens. \citet{goyal2024thinkspeaktraininglanguage} similarly find that learnable pause tokens help reasoning models, but the benefit does not transfer to standard pretrained models at test time. \citet{greenblatt2024counting} shows that recent frontier models can leverage filler tokens without any additional training, including tokens that repeat the question, suggesting this is an emergent ability~\citep{wei2022emergentabilitieslargelanguage}. \citet{shi2025meaninglesstokensmeaningfulgains} attribute such gains to shifts in MLP activations and reproduce them with an inference-time intervention. Relative to these works, we characterize how the uplift depends jointly on filler token type, few-shot context, task, and model, and we connect filler tokens to hidden constraint satisfaction and training dynamics. \citet{balesni2025lessonsstudyingtwohop} further demonstrate that frontier models are capable of performing multi-hop invisible reasoning without chain of thought.

\paragraph{Deception, Scheming, and CoT Monitoring}
A growing body of work shows that language models can pursue hidden objectives. \citet{hubinger2024sleeperagents} demonstrate that deceptive behaviors can be implanted and persist through safety training. \citet{greenblatt2024alignmentfaking} show that models can infer when they are being evaluated and strategically comply with harmful requests to preserve their values during post-training, and \citet{meinke2025frontiermodelscapableincontext} show that this extends to in-context scheming without requiring training. \citet{zolkowski2025reasoningmodelsobfuscatereasoning} show that under optimization pressure, models can complete adversarial tasks while hiding their intent from CoT monitors, and \citet{baker2025monitoringreasoningmodels} find that optimizing against a CoT monitor causes models to conceal reward hacking within their reasoning. \citet{marks2025auditinghiddenobjectives} show that interpretability tools can recover some hidden objectives that CoT monitoring misses, and \citet{fang2025unsuperviseddecoding} show that logit-lens analysis can reconstruct hidden reasoning when a model is trained to reason in a cipher.

\section{Methodology}
\label{sec:definitions}

Language models do not always express their reasoning through an interpretable chain-of-thought. We define \textit{invisible reasoning} as consequential computation that occurs in internal latent representations during a model's forward pass without leaving any interpretable trace in the output tokens. Inserting semantically irrelevant tokens in the model's context can strengthen invisible reasoning by scaling up forward pass computation~\citep{pfau2024letsthinkdotdot}. We refer to these irrelevant tokens as \textbf{filler tokens}. These tokens are irrelevant in the sense that the same fixed sequence appears for every question, so the tokens carry no information about any particular problem or answer. The tokens may still act as procedural cues, and our second criterion below directly measures this possibility.

We identify three kinds of invisible reasoning phenomena: 
\begin{enumerate}
    \item \textbf{Performance improves with filler tokens}, indicating the presence of invisible reasoning.
    \item \textbf{Performance depends on filler token content}, suggesting that certain tokens' representations are more favorable than others.
    \item \textbf{Filler token preferences vary across models}, demonstrating that the filler token uplift is not due to semantic content but rather the model-specific token representations.
\end{enumerate}

These criteria establish that filler tokens causally affect a model's latent computation, but it is unclear what algorithm a model executes in the filler span. Inserting tokens also changes positions and attention patterns, so we do not attribute uplift to extra computation alone. In Section~\ref{sec:mechanistic} we probe the underlying mechanism behind latent computation in the presence of filler tokens.

\subsection{Synthetic Tasks}
The goal of this paper is to isolate and quantify the existence of invisible reasoning.  We focus on simple synthetic test scenarios for which the impact of reasoning can be isolated and quantified.  We also focus on questions with verifiable true/false answers to remove confounding variables and subjectivity. 
We additionally choose tasks where the problem statement is terse, so that prefilled filler tokens meaningfully increase the amount of test-time computation. We provide complete task details and seeds for reproducibility in Appendix~\ref{app:task_details}.

\paragraph{Multi-Digit Multiplication.}  We generate a synthetic dataset of multi-digit integer multiplication problems, using 4-digit operands in our main experiments. Each problem is of the form \texttt{What is X times Y?}

\paragraph{Multi-Step Arithmetic.} Following~\citet{greenblatt2024counting} we construct a synthetic dataset where each problem consists of 5 to 7 nested arithmetic operations. 

\paragraph{Variable Counting in Code.} We provide the model with a short code snippet and ask it to output how many distinct variables are assigned a value in the snippet. 

Later, we will consider slightly more complex setups in which the model must additionally fulfill a hidden mathematical goal while ostensibly solving one of the above tasks.

\subsection{Setup}
We design an evaluation framework to probe for the existence of invisible reasoning based on our three proposed criteria. Given a task, we prompt the model to answer immediately without producing a CoT. We include $K$ few-shot examples, where each example consists of a question, $n$ filler tokens in the assistant context, and the ground-truth answer. We also prefill $n$ filler tokens in the assistant context before prompting the model to generate the answer for each question. Throughout the paper, $n$ denotes the number of filler tokens and $N$ the number of evaluation problems. We measure baseline performance with the same setup, except we do not include filler tokens in the few-shot examples or after the question. We provide additional prompting details and ablations in Appendix~\ref{app:prompts}.

We experiment with 17 filler token types, which span numerical sequences, word lists, and non-semantic symbols (Section~\ref{sec:filler_type}). Each type maps to a fixed sequence, such as counting numbers in order or a fixed list of animal names, truncated to the target length; random number and random token sequences are also fixed across problems. We calibrate each sequence to contain approximately $n$ tokens under the target model's tokenizer (Appendix~\ref{app:calibration}). We sample using temperature 1 with top-$p = 0.95$, and we prevent reasoning for API models by disabling the reasoning parameter on OpenRouter. For each evaluation we report sample mean accuracy with 95\% confidence intervals computed from the standard error. All settings within a task share the same 1{,}000 problems, so comparisons between settings are paired. Because we sweep many filler types, models, and prompting settings, we emphasize effects that replicate across settings over any single comparison. Unless stated otherwise, we prefill filler tokens ourselves; when a model instead generates its own filler tokens, we say so explicitly, since generated and prefilled filler tokens yield different selection effects. 
To prevent models from producing explicit reasoning, we prefill the ``Answer:'' prefix in the assistant context to the model to generate the answer directly, as detailed 
in Appendix~\ref{app:calibration}.

\section{Invisible Reasoning with Filler Tokens}
\label{sec:filler}

We analyze how Qwen3-235B-Instruct-2507 leverages invisible reasoning to improve performance on our three synthetic tasks. We first sweep across 17 filler token types in Section~\ref{sec:filler_type}. In Figure~\ref{fig:filler_tasks} we investigate how filler token uplift varies with different tasks and few-shot prompting setups. We conclude with a mechanistic analysis in Section~\ref{sec:mechanistic}, showing that invisible reasoning is distributed across the full filler sequence, order-sensitive, and established in early layers of the model. Across all of our results, \textbf{filler tokens enable models to beat a no-CoT baseline} despite carrying no information about the problem being solved.

\subsection{Sensitivity to Filler Token Choices}
\label{sec:filler_type}
\begin{figure}[t]
\centering
\includegraphics[width=0.8\linewidth]{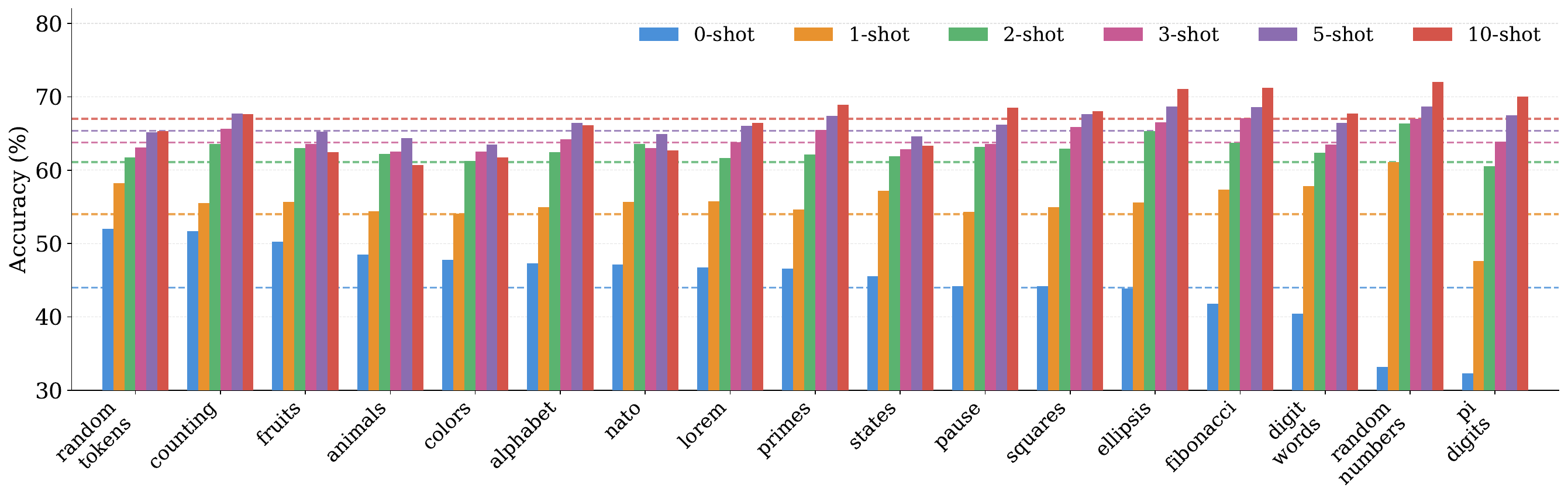}
\caption{\textbf{Invisible reasoning uplift depends on filler token type and few-shot prompting.} We evaluate Qwen3-235B on 4-digit multiplication with 17 different types of filler tokens, sorting by zero-shot accuracy improvement. Dashed lines indicate baseline accuracy for each few-shot setting. Invisible reasoning uplift significantly depends on the type of filler token. Notably, filler token types that harm zero-shot accuracy the most (right side) yield the best performance in the 5-shot and 10-shot settings, and vice versa.}
\label{fig:filler_type_fewshot}
\end{figure}

In Figure~\ref{fig:filler_type_fewshot} we illustrate filler token evaluation results across all filler token types and multiple few-shot settings. We provide examples of each filler token sequence in Appendix~\ref{app:filler_examples}. Many filler token types, such as random numbers and pi digits, are harmful in the zero-shot setting. However, these same tokens provide accuracy uplift when prefilled alongside 10 few-shot examples. Appendix~\ref{app:cv} (Table~\ref{tab:qwen_sweep_summary}) confirms this dependence across filler token counts and few-shot settings. Figure~\ref{fig:filler_type_fewshot} demonstrates a \textbf{type inversion}, where the most beneficial filler token types in the zero-shot setting are harmful in 10-shot, and vice versa. In all few-shot settings, we observe a strong dependence between performance and filler token type, which we verify with a statistical analysis in Appendix~\ref{app:cv}. The inversion shows that filler token uplift depends on the token type and the preceding context jointly. This finding indicates that the attention mechanism plays a key role in extracting useful information from filler token representations.

\begin{figure}[t]
\centering
\includegraphics[width=0.8\linewidth]{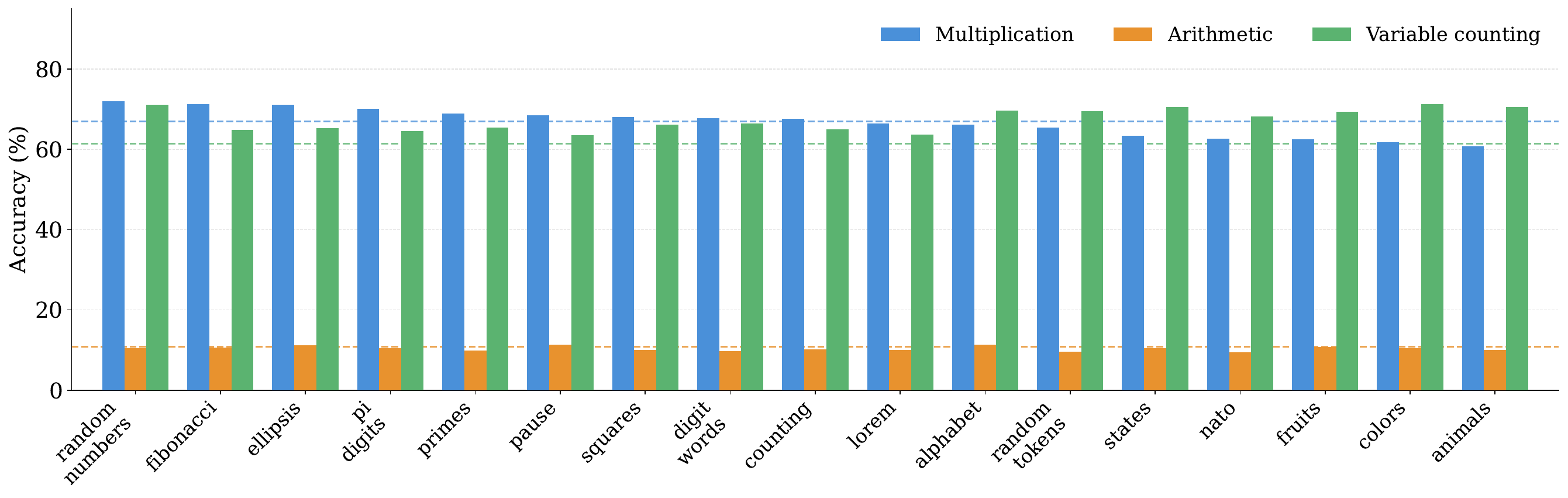}
\caption{\textbf{Invisible reasoning uplift varies across tasks.} Dashed lines denote baseline performance for each task, with filler types sorted by uplift on 4-digit multiplication. Qwen3-235B achieves very low baseline accuracy on the multi-step arithmetic task, resulting in no significant filler token uplift. On our variable counting task, all filler token types provide uplift compared to the baseline. However, on the multiplication task, uplift depends on the filler token type as seen in Figure~\ref{fig:filler_type_fewshot}.}
\label{fig:filler_tasks}
\end{figure}

We now evaluate Qwen3-235B in the 10-shot setting on all three synthetic tasks in Figure~\ref{fig:filler_tasks}. For a given filler token type, the performance uplift relative to the baseline depends on the task. For the arithmetic task, no filler type yields uplift, and on the variable counting task all filler types provide positive performance gain. On the multiplication task, however, performance uplift is sensitive to the filler token type. \textbf{The effect of invisible reasoning depends jointly on the filler tokens, the prompting scheme, and the specific task.}

We also evaluate Qwen3.5-397B~\citep{qwen35blog} and find that it is highly sensitive to filler token content. On arithmetic, the model achieves near $0\%$ accuracy regardless of filler token type, and filler tokens provide no benefit. On multiplication, the model achieves $90.6\%$ at 10-shot when served locally (Appendix~\ref{app:qwen35_sweep}), and filler token effects are extreme: Fibonacci tokens improve accuracy by $+1.5\%$, while word-based types (animals, NATO, fruits, colors) drop accuracy to near $0\%$. These drops coincide with format violations, where the model abandons its concise answer format and produces long outputs. Some filler tokens affect outputs by shifting the model's behavioral mode rather than through latent computation alone. Full results are in Appendix~\ref{app:qwen35_sweep}. Although invisible reasoning cannot match the performance of standard CoT, we observe meaningful uplift over the zero-CoT baseline despite the filler tokens carrying no semantic relevance to the task.

\subsection{Mechanistic Analysis}
\label{sec:mechanistic}
We now ask which internal mechanisms drive the filler token uplift in open-weight models. The type inversion across few-shot settings in Figure~\ref{fig:filler_type_fewshot} suggests that filler tokens interact with the surrounding context in non-trivial ways. To investigate this, we run three mechanistic experiments on Qwen3-235B on the variable counting task, comparing animals filler tokens (which provide strong uplift) against Fibonacci tokens (which provide weak uplift). Full details are in Appendix~\ref{app:mechinterp}.  Interestingly, we observe striking differences in behavior between the filler token types.

\paragraph{Filler token order.} We construct mixed filler sequences by concatenating 50 animals tokens followed by 50 Fibonacci tokens, and vice versa. The ``animals$\to$Fibonacci'' ordering recovers most of the uplift from pure animals tokens, while ``Fibonacci$\to$animals'' performs \emph{worse} than pure Fibonacci. The asymmetry suggests that early positions in the filler sequence are disproportionately important compared to later tokens.

\paragraph{Activation patching.} We patch residual stream activations~\citep{zhang2024towards} from animals filler tokens into a forward pass that otherwise uses Fibonacci tokens, applying the patch across all 94 layers of Qwen3-235B. Patching at layers 0--30 recovers over 90\% of the performance gap between the two filler types; patching at layers 70 and above yields negative recovery, indicating that late-layer activations from animals tokens are harmful when inserted into a Fibonacci context. Patching a single token position is insufficient, suggesting that the useful signal is distributed across the full filler sequence and originates in early layers. We note that early-layer activations may steer later computation without themselves containing the useful algorithmic state.

\paragraph{Linear probing.} We train linear probes on mean-pooled residual stream activations across the filler span at each layer to predict the correct answer. The probe accuracy for animals filler tokens exceeds both Fibonacci and the no-filler baseline from layer 15 onwards, suggesting that task-relevant information accumulates in filler token representations throughout the forward pass, and that the choice of filler token type determines how effectively this accumulation occurs. Probe accuracy shows that this information can be decoded from the filler span, though the model may not necessarily use it.

Across all three experiments, \textbf{the filler sequence as a whole, rather than any single token position, appears to carry the useful signal}, and the ordering of tokens within the span matters. The benefit emerges in early layers, consistent with filler tokens shaping residual stream activations early in the forward pass. We describe additional attention masking experiments in Appendix~\ref{app:attention_masking}; these do not reveal a single universal mechanism across tasks and filler token types. Notably, on zero-shot multiplication the uplift survives even when the filler span cannot attend to any prior context (Table~\ref{tab:mask_core}). In this setting the benefit cannot come from computation specific to the question, so the mechanism may instead be related to attention redistribution or positional effects. Practically, the probing results suggest that activation-level monitors could detect filler-based computation that token-level monitors miss. Having established that frontier models can perform invisible reasoning, we now ask how broadly this capability extends across the frontier model landscape.

\begin{figure}[t]
\centering
\includegraphics[width=0.8\linewidth]{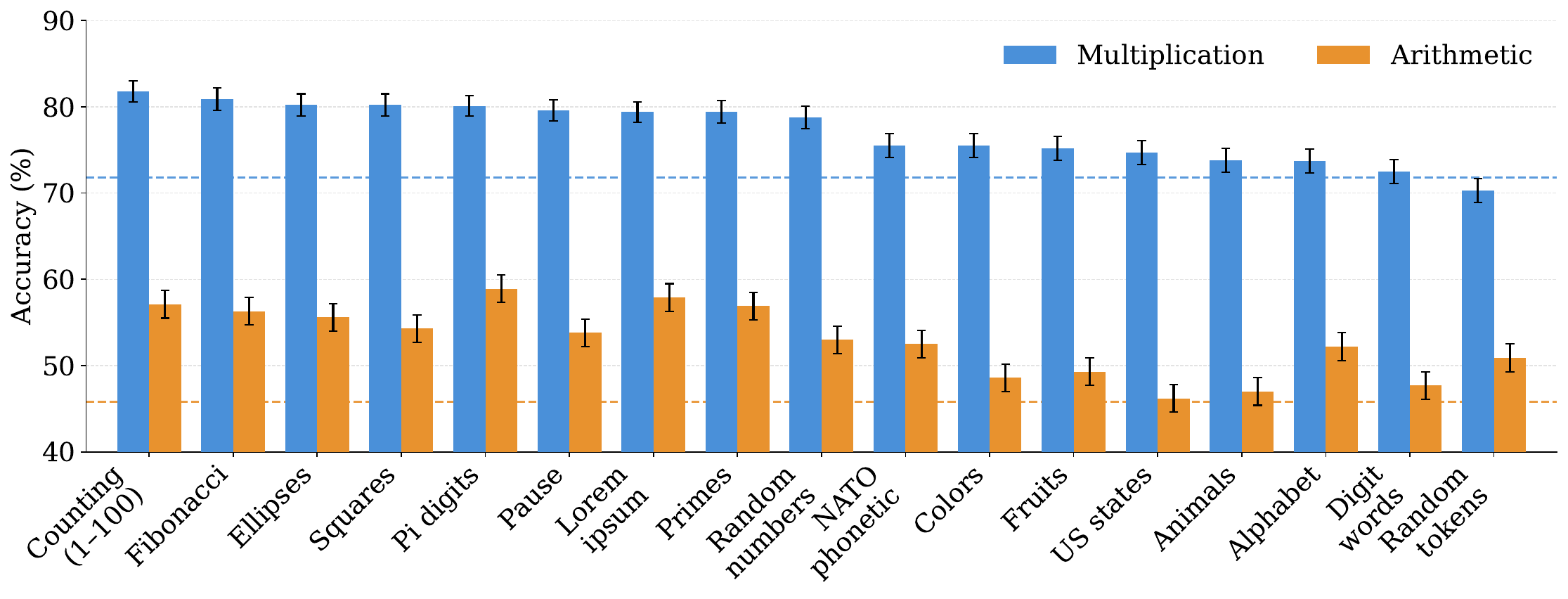}
\caption{\textbf{Filler token uplift varies across tasks for Claude Opus~4.5}. We compare performance across 17 filler token types in the 10-shot setting for both 4-digit multiplication and multi-step arithmetic. Filler token uplift depends on the type of token, but whether a token provides uplift also depends on the specific task. Even for the same task and token type, performance uplift varies significantly between Claude Opus~4.5 and Qwen3-235B.}
\label{fig:filler_types}
\label{tab:filler_types}
\end{figure}

\section{Frontier Model Evaluation}
\label{sec:frontier}

We now investigate whether filler token performance varies across language models. We evaluate the following frontier language models: Claude Opus 4.5, Claude Opus 4.6, Claude Sonnet 4.5, Gemini 3 Flash, Kimi K2.5~\citep{kimiteam2026kimik25visualagentic}, DeepSeek V3.2~\citep{deepseekai2025deepseekv32pushingfrontieropen}, Llama 4 Maverick, Qwen3-235B-A22B Instruct-2507~\citep{yang2025qwen3technicalreport}, Qwen3.5-397B~\citep{qwen35blog}, GLM 4.7~\citep{glm2024chatglmfamilylargelanguage}, GLM 5~\citep{glm5}, GPT-5.2, and GPT-5.5. All models are evaluated via API access through OpenRouter. Opus~4.6, GPT-5.2, and GPT-5.5 do not support assistant prefilling, so their results may carry selection effects (Appendix~\ref{app:calibration}). Moreover, we cannot control each provider's serving configuration. We therefore present Table~\ref{tab:cross_model} as evidence of within-model filler effects rather than a ranking of model performance.

\subsection{Filler Token Performance Varies Across Models}
We begin by evaluating Claude Opus~4.5 in the 10-shot setting on both 4-digit multiplication and multi-step arithmetic. Figure~\ref{fig:filler_types} illustrates the results. When comparing performance between multiplication and arithmetic, we see that filler token performance depends on the task. For example, random tokens do not improve performance for multiplication, but they do yield uplift for arithmetic. Additionally, the token hierarchy differs significantly from Section~\ref{sec:filler}: counting tokens provide the greatest uplift for Claude Opus~4.5, but they do not yield any improvement for  Qwen3-235B in the 10-shot setting. We note that invisible reasoning is more robust in Claude Opus~4.5 compared to Qwen3-235B; almost all filler token types provide some uplift, and no filler tokens harm accuracy relative to the baseline. We omit variable counting from Figure~\ref{fig:filler_types} and Table~\ref{tab:cross_model} because Opus~4.5 and most other frontier models saturate the task; Qwen3-235B does not, and its results on this task appear in Figure~\ref{fig:filler_tasks}.

\subsection{Cross-Model Comparison}
In evaluating both Claude Opus~4.5 and Qwen3-235B, we see that invisible reasoning with filler tokens is sensitive to the type of filler token, the task, and the prompting scheme. We now test for invisible reasoning across thirteen frontier models using counting filler tokens in Table~\ref{tab:cross_model}. Many existing frontier models, such as Claude Opus and Gemini 3, benefit significantly from filler tokens. We again observe the effects of task dependence; Gemini 3 receives $>10\%$ uplift from filler tokens on the arithmetic task, but no statistically significant improvement on 4-digit multiplication. We note that Qwen3-235B demonstrates no uplift in this evaluation as it is specific to counting tokens; however, our results in Section~\ref{sec:filler_type} show that this does not rule out possible invisible reasoning capability with other types of filler tokens.

\begin{table}[t]
\centering
\caption{\textbf{Invisible reasoning capability varies significantly across models}. We evaluate each model in the 10-shot setting using counting tokens (1--100), with $\Delta$ denoting the change in accuracy relative to the 10-shot baseline without filler tokens. Many frontier models can leverage filler tokens to improve accuracy, with Gemini 3 and Claude Opus showing the strongest uplift. GPT-5.5 nearly saturates both tasks in the baseline setting, but filler tokens close the remaining gap on multiplication ($98.5\% \to 100\%$). GLM-5 shows no statistically significant improvement on either task. Asterisks mark models whose APIs do not permit assistant prefilling, so their results may carry selection effects (Appendix~\ref{app:calibration}).}
\vspace{0.05in}
\label{tab:cross_model}
\small
\rowcolors{2}{gray!15}{white}
\begin{tabular}{@{}lcccccc@{}}
\toprule
& \multicolumn{3}{c}{\textbf{Arithmetic}} & \multicolumn{3}{c}{\textbf{4-digit Multiplication}} \\
\cmidrule(lr){2-4} \cmidrule(lr){5-7}
\textbf{Model} & \textbf{Baseline} & \textbf{1--100} & $\boldsymbol{\Delta}$ & \textbf{Baseline} & \textbf{1--100} & $\boldsymbol{\Delta}$ \\
\midrule

\multicolumn{7}{@{}l}{\textbf{Closed-source models}} \\
\midrule

Opus 4.6$^{*}$ & $61.7_{\pm 2.4}$ & $91.7_{\pm 3.1}$ & $+30.0$ & $65.7_{\pm 1.5}$ & $72.8_{\pm 1.4}$ & $+7.1$ \\
Gemini 3 Flash & $51.0_{\pm 1.7}$ & $61.7_{\pm 1.6}$      & $+10.7$ & $96.0_{\pm 0.6}$ & $97.0_{\pm 0.6}$ & $+1.0$ \\
Opus 4.5       & $45.8_{\pm 1.6}$ & $57.0_{\pm 1.6}$      & $+11.2$ & $71.8_{\pm 1.4}$ & $81.8_{\pm 1.2}$ & $+10.0$ \\
Sonnet 4.5     & $31.5_{\pm 1.5}$ & $40.2_{\pm 1.6}$      & $+8.7$  & $61.0_{\pm 1.5}$ & $65.8_{\pm 1.5}$ & $+4.8$ \\
GPT-5.2$^{*}$  & $23.2_{\pm 1.3}$ & $26.2_{\pm 1.9}$ & $+3.0$  & $33.2_{\pm 1.5}$ & $30.9_{\pm 1.4}$ & $-2.3$ \\
GPT-5.5$^{*}$  & $99.8_{\pm 0.1}$ & $99.9_{\pm 0.1}$ & $+0.1$  & $98.5_{\pm 0.4}$ & $100.0_{\pm 0.0}$ & $+1.5$ \\
\midrule

\multicolumn{7}{@{}l}{\textbf{Open-weight models}} \\
\midrule

DeepSeek V3.2    & $34.5_{\pm 1.5}$ & $32.7_{\pm 1.5}$ & $-1.8$ & $70.6_{\pm 1.4}$ & $77.9_{\pm 1.4}$ & $+7.3$ \\
Kimi K2.5        & $23.4_{\pm 1.4}$ & $22.8_{\pm 1.4}$ & $-0.6$ & $92.0_{\pm 0.9}$ & $95.1_{\pm 0.7}$ & $+3.1$ \\
Qwen3.5-397B     & $18.9_{\pm 1.2}$ & $16.0_{\pm 1.3}$ & $-2.9$ & $99.3_{\pm 0.3}$ & $99.2_{\pm 0.3}$ & $-0.1$ \\
Llama 4 Maverick & $13.6_{\pm 1.1}$ & $14.1_{\pm 1.1}$ & $+0.6$ & $77.0_{\pm 1.3}$ & $82.7_{\pm 1.2}$ & $+5.7$ \\
GLM-4.7          & $12.4_{\pm 1.0}$ & $12.6_{\pm 1.0}$ & $+0.2$ & $31.5_{\pm 1.5}$ & $35.8_{\pm 1.5}$ & $+4.3$ \\
GLM-5            & $11.9_{\pm 1.0}$ & $13.4_{\pm 1.1}$ & $+1.5$ & $42.2_{\pm 1.6}$ & $39.2_{\pm 1.5}$ & $-3.0$ \\
Qwen3-235B       & $11.0_{\pm 1.0}$ & $9.7_{\pm 0.9}$  & $-1.3$ & $69.6_{\pm 0.5}$ & $69.2_{\pm 0.5}$ & $-0.4$ \\
\bottomrule
\end{tabular}
\end{table}

\subsection{Satisfying Hidden Constraints}
\label{sec:hidden_motive}

We test whether filler tokens help models secretly pursue hidden goals via invisible reasoning. In this setting, the system prompt for the multiplication task instructs Claude Opus~4.5 and Qwen3-235B to output N/A instead of the answer to the user prompt when the true answer satisfies one of three hidden modular constraints of increasing complexity: $x \bmod 2 = 1$, $x \bmod 3 = 2$, or $4x \bmod 7 = 5$. Satisfying the hidden constraint requires the model to compute a divisibility property of an answer it has not yet produced, and no token in the context or output reveals whether that computation occurred. Filler tokens improve the accuracy of this unobservable step.

Table~\ref{tab:hidden_motive} reports N/A accuracy over samples satisfying the constraint. For Opus~4.5, filler tokens improve the easy constraint from $33.5\%$ to $44.5\%$, with no significant gains on harder constraints. For Qwen3-235B, filler tokens yield $98.8\%$ accuracy on the easy constraint and $39.0\%$ on the medium constraint. This is notable because counting filler tokens do not improve Qwen's multiplication accuracy alone in this 10-shot setting~(Fig.~\ref{fig:filler_type_fewshot}), yet they improve hidden-goal satisfaction.

Opus~4.5 has the same multiplication accuracy on non-N/A samples in this setting as it does in Table~\ref{tab:cross_model}. In contrast, Qwen3-235B's multiplication accuracy drops substantially on non-N/A samples, suggesting that its large hidden-goal gains come from learning a modular shortcut at the expense of the primary task. The strongest form of this result requires hidden goal uplift without sacrificing primary task accuracy; Opus~4.5 meets this criteria on the easiest constraint, while Qwen3-235B does not. Full Qwen3-235B results appear in Appendix~\ref{app:hidden_motive_qwen}.

\begin{table}[htbp]
\centering
\caption{\textbf{Filler tokens improve hidden goal accuracy.} Filler tokens enable models to satisfy a hidden modular arithmetic constraint alongside a primary multiplication task. Opus~4.5 achieves uplift on the easiest hidden goals while preserving multiplication accuracy on non-N/A samples. Qwen3-235B achieves much higher hidden goal satisfaction, but this comes at the cost of accuracy on non-NA samples (Appendix~\ref{app:hidden_motive_qwen}). Baseline~(0-CoT) uses the same 10-shot prompt with no CoT and no filler tokens; the gap between the two columns for each model is the filler token uplift in hidden goal accuracy.}
\label{tab:hidden_motive}
\small
\begin{tabular*}{\linewidth}{@{\extracolsep{\fill}}lcccc@{}}
\toprule
& \multicolumn{2}{c}{\textbf{Opus~4.5}} & \multicolumn{2}{c}{\textbf{Qwen3-235B}} \\
\cmidrule(lr){2-3} \cmidrule(lr){4-5}
\textbf{Hidden Goal} & \textbf{Baseline~(0-CoT)} & \textbf{1--100 Filler Tokens} & \textbf{Baseline~(0-CoT)} & \textbf{1--100 Filler Tokens} \\
\midrule
$4x \bmod 7 = 5$ & $0.3\%$ & $1.2\%$ & $0.0\%$ & $0.0\%$ \\
$x \bmod 3 = 2$ & $8.7\%$ & $10.0\%$ & $0.0\%$ & $39.0\%$ \\
$x \bmod 2 = 1$ & $33.5\%$ & $44.5\%$ & $0.0\%$ & $98.8\%$ \\
\bottomrule
\end{tabular*}
\end{table}

\section{What Training Methodologies Elicit Invisible Reasoning?}
\label{sec:rl}
We choose Qwen3-235B, an open-weight model that only displays invisible reasoning in a narrow range of settings (Section~\ref{sec:filler}), to study whether invisible reasoning manifests in different forms or at higher intensities under optimization pressure. SFT fails to transfer invisible reasoning across all configurations we tested; full details are in Appendix~\ref{app:sft}. Our RL findings are also largely negative: RL reshapes filler token preferences and improves the underlying task, but it does not produce a filler token advantage that persists at test time (Appendix~\ref{app:rl_results}).

\noindent\textbf{RL setup.}
We RL all parameters of Qwen3-235B-A22B-Instruct-2507 with a PPO-style clipped policy gradient objective~\citep{schulman2017ppo} using GRPO-style~\citep{shao2024deepseekmath} group advantage estimation with MaxRL normalization~\citep{tajwar2026maximumlikelihoodreinforcementlearning}.
We compute advantages only over answer tokens, and we do not compute any reward signal related to the filler tokens. The filler token distribution is still optimized indirectly, since answer tokens' gradients backpropagate through the hidden states that filler tokens create.
Full details of the objective, including the hyperparameters, prompts, clipping bounds, hard ratio masking, KL monitoring, and infrastructure (asynchronous RL, rollout router replay for the MoE architecture), are in Appendix~\ref{app:training_details}.
All of the prompts in our RL experiments use 10 few-shot examples. Similar to the previous experiments, each few-shot example includes a question and the correct answer with zero CoT~(for baselines), or filler tokens~(for models of interest). We vary the filler type in the following experiments.
We run all RL experiments on H100 GPUs using \href{https://github.com/togethercomputer/xorl}{an open-source RL codebase.}

\subsection{Emergent Filler Token Preferences During RL}
\begin{figure}[t]
    \centering
    \includegraphics[width=0.45\linewidth]{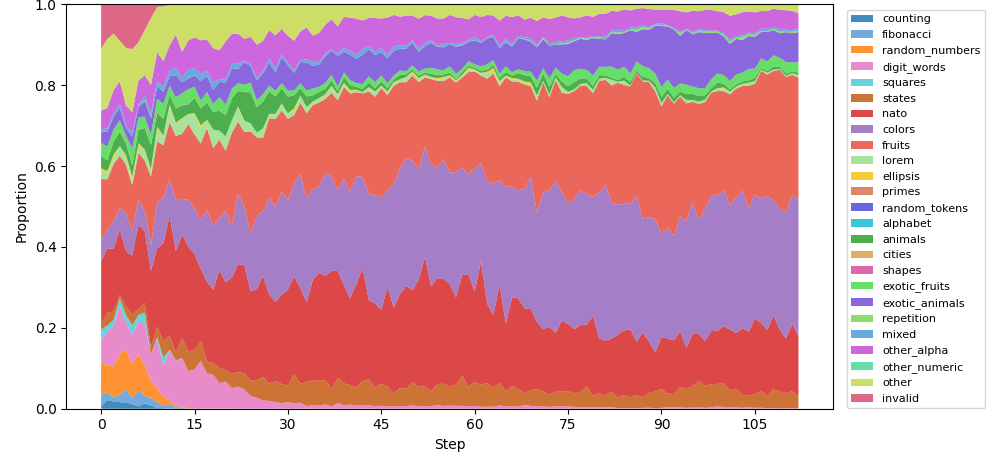}
    \hfill
    \includegraphics[width=0.45\linewidth]{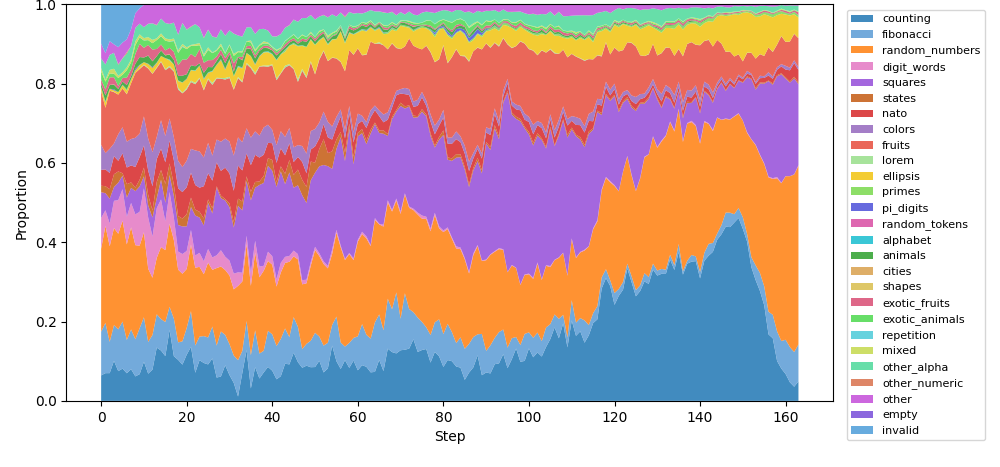}
    \caption{\textbf{Learned filler token preferences during RL depend on the task.} Qwen3-235B quickly develops strong preferences for select filler token types on 4-digit multiplication (left), but preferences on multi-step arithmetic (right) continue to shift after many training steps. In both settings, the model quickly assigns $\approx 0$ probability to several filler tokens before continuing to explore a smaller subset of types throughout the rest of training.}
    \label{fig:strategy-proportions}
\end{figure}

We provide 10 few-shot examples whose filler token types are sampled uniformly from the full set in Figure~\ref{fig:filler_type_fewshot}, and train Qwen3-235B with RL on both 4-digit multiplication and multi-step arithmetic.
Figure~\ref{fig:strategy-proportions} charts the proportion of canonical filler types over training, along with novel sequences the model invents (cities, shapes, repeating digits, exotic animal and fruit names). The two tasks exhibit qualitatively different dynamics. On multiplication~(left), strategy diversity collapses early in training. The model settles on NATO, colors, and fruits after around 30 steps, and the benefit of self-generated filler over 10-shot no-filler sampling is marginal ($+1.07\%$ pass@8). On arithmetic~(right), the model continues shifting strategies late in training, but the final improvement over no-filler is not substantially larger. Full reward and KL curves are in Appendix~\ref{app:rl_results}.

\noindent \textbf{What is RL doing?} RL training improves pass@8 but yields little gain in pass@1, suggesting that filler tokens primarily affect the \textit{diversity} of model outputs rather than the accuracy of any single forward pass. Greater diversity across independently sampled rollouts is enough to explain this pattern. We speculate that filler tokens allow the model to internally explore multiple candidate answers within a single forward pass, with the prefilled span acting as a workspace over which different latent computations can be conditioned. Under this view, RL is not teaching the model to increase accuracy directly, but rather to allocate its internal capacity across a wider set of candidate solutions. This interpretation is speculative, and we do not provide direct evidence that distinct candidate answers are simultaneously represented in the residual stream. If the interpretation holds, \textbf{models may use filler tokens as an internal search mechanism}. Testing this hypothesis directly, for example by resampling answers from the cached filler states, is a natural next step for future work.

\subsection{Hidden Computation in Natural Reasoning Traces}
\label{sec:gsm8k}
We now ask whether a model can instead encode hidden computation within a semantically meaningful CoT; specifically, we test whether reasoning about one question can carry information useful for answering a second question. Across all reward functions we tried, the answer is largely negative: the model consistently prefers reward hacking over encoding hidden information.
The model now generates its own tokens rather than being supplied fillers, and those tokens contain reasoning about $q_1$ rather than being semantically empty. We train Qwen3-235B on GSM8k~\citep{cobbe2021gsm8k}. In the dual-question setting, the model receives questions $q_1$ and $q_2$, produces traces $r_1$ and $r_2$, and is evaluated on accuracy for $q_2$. Both questions appear in a single prompt, and the model produces $r_1$, $r_2$, and both answers in one generation. We use the single-question setting for the antidistillation objective below, and the dual-question setting for the monitored reasoning and length penalty objectives. We now apply RL with various reward functions and use trace ablations (removing or replacing $r_1$ with random tokens) to determine whether $q_2$ accuracy depends on the content of $r_1$.

\noindent \textbf{Antidistillation} The reward $R = \mathbbm{1}[\hat{y} = y] - \Pr[\hat{y}_{\text{base}} = y \mid r]$ incentivizes the model to produce traces that mislead a base model while retaining accuracy. The model discovers two obfuscation strategies. In the first, the model states the correct answer early in its CoT, introduces a cue word such as ``wait'' or ``however,'' then appends spurious reasoning that leads to an incorrect conclusion. The trained model uses the cue word to retrieve the first answer; the base model is misled by the full CoT. In the second, the model expresses false uncertainty (``there is not enough information''), causing the base model to lose confidence while the trained model ignores its own expressed doubt. 
Backtracking tokens like ``wait'' or ``however'' are common in existing frontier model reasoning traces, suggesting a potential mechanism by which models could exploit visible reasoning for hidden computation.

\noindent \textbf{Monitored Reasoning} GPT-4o queried via API serves as the monitor, rating $q_2$-relevant content in $r_1$ from 1 to 10; we divide the rating by 10 to obtain $s_{\text{monitor}}$. When the monitor penalizes $q_2$-relevant content in $r_1$ with reward $R = \mathbbm{1}[\hat{y}_2 = y_2] - 0.1 \cdot s_{\text{monitor}}$, the model achieves $96.5_{\pm 0.3}\%$ $q_2$ accuracy by producing vacuous traces that evade the monitor. However, replacing $r_1$ with a single random token fully restores accuracy after removing $r_1$ degrades it to $94.4_{\pm 0.4}\%$, indicating the model relies only on the presence of a token in the $r_1$ position as a cue rather than its content.

\noindent \textbf{Length Penalty} With reward $R = \mathbbm{1}[\hat{y}_2 = y_2] - |r_2| / (|r_1| + |r_2|)$, the model is incentivized to shift its reasoning from $r_2$ to $r_1$. After a monitor is added, the model instead inflates $r_1$ with lengthy $q_1$ reasoning to reduce the length penalty fraction. Our trace ablation confirms that removing $r_1$ does not degrade $q_2$ accuracy, while removing $r_2$ does. The model never compresses $q_2$ reasoning into $r_1$ because it can achieve arbitrarily high reward by lengthening $r_1$ instead.

In the dual-question setting, the model never encodes $q_2$-relevant information within $r_1$, instead learning reward hacks in all conditions. In the single-question antidistillation setting, however, the model does encode surface-level information in its own reasoning trace via backtracking cues like ``wait'' or ``however.'' This is a weaker form of hidden computation than invisible reasoning, but demonstrates that models can exploit visible tokens as private signals even without filler tokens.

\paragraph{Limitations.} Our tasks are synthetic, and the filler token uplift is small relative to the gains from explicit CoT on the same problems. The mechanistic and RL analyses are limited to a single model, Qwen3-235B. Our hidden constraint experiments demonstrate a capability rather than monitor evasion in a realistic misbehavior setting. Our attention masking results also show that uplift can persist when the filler span cannot attend to the question, so mechanisms other than question-specific computation contribute in some settings. Finally, we do not evaluate fillers that repeat the question~\citep{greenblatt2024counting} or natural CoT text used as filler; both are natural extensions of our experiments.

\section{Conclusion}
As future frontier models grow larger and more capable, we expect invisible reasoning to become more robust, since greater internal computation provides more capacity for hidden processing. Our results show that \textbf{frontier models already perform consequential computation in tokens with no semantic meaning}. Our findings motivate several future research directions: developing interpretability tools that audit residual stream activations during filler spans, designing RL objectives that explicitly probe for latent computation, and constructing monitoring schemes that remain robust when models can offload reasoning into latent representations.

Our paper studies invisible reasoning directly, and our RL experiments attempt to elicit this behavior from models. We believe this work remains aligned with the broader incentives of RL research. Invisible reasoning has a clear practical motivation: prefill is typically compute-bound while decode is memory-bound, and current accelerators continue to improve TFLOPS faster than memory bandwidth. A model that shifts test-time compute from decode into prefill could therefore serve responses more cheaply, though realized savings depend on batching, context length, and serving configuration. Recent work explores training directions motivated by these efficiency considerations: \citet{ramji2026thinkingwordsefficientlatent} pursue an abstract CoT objective that would also invalidate CoT monitoring, and \citet{geiping2026scaling} study latent reasoning via looping transformers. We expect frontier labs to pursue training objectives aligned with these incentives. We therefore consider it imperative to study invisible reasoning openly. Invisible reasoning already exists in frontier models, and the field of AI safety research will need to develop monitoring schemes to address this behavior.

\section*{Acknowledgements}
This work was supported by a grant from Coefficient Giving. Compute for initial experiments was provided by a Tinker Academic Research Grant. Later experiments were supported by TogetherAI. Vatsal is funded by an NSF GRFP. Ashwinee was supported by the Constellation Visiting Fellows Program. We would like to thank researchers from Redwood Research (Ryan Greenblatt and Andy Wang) and Anthropic (Fabien Rogers and Matthew Jagielski) for feedback that was instrumental from brainstorming through publication.

This material is based upon work supported by the National Science Foundation Graduate Research Fellowship Program under Grant No. DGE-2234660. Any opinions, findings, and conclusions or recommendations expressed in this material are those of the author(s) and do not necessarily reflect the views of the National Science Foundation.

\bibliography{main}
\bibliographystyle{plainnat}

\appendix

\section{Invisible Reasoning Evaluation Details}
\label{app:eval_details}

\subsection{Coefficient of Variation Analysis}
\label{app:cv}

The coefficient of variation (CV $=$ Std$/|$Mean$|$ of accuracy uplift across filler types) provides a direct measure of type dependence. Low CV indicates that all filler types produce similar uplift, consistent with additional forward pass computation being the operative factor. High CV indicates that uplift varies substantially across types, meaning the specific token representations matter. CV becomes unstable when the mean uplift is near zero (e.g., CV${=}7.4$ at $n{=}100$, 20-shot), so we only interpret CV in settings where the mean uplift is clearly separated from zero. Table~\ref{tab:qwen_sweep_summary} reports CV for Qwen3-235B on 4-digit multiplication across token counts and few-shot settings.

\begin{table}[h]
\centering
\caption{\textbf{CV analysis for Qwen3-235B on 4-digit multiplication.} Mean and Std of accuracy change ($\Delta$) are across all 17 filler types. CV $=$ Std/$|$Mean$|$: low CV indicates type-independent uplift, high CV indicates type-dependent uplift. CV is unstable when Mean $\Delta$ is near zero. \#pos and \#neg denote how many filler types improve or hurt accuracy.}
\label{tab:qwen_sweep_summary}
\small
\rowcolors{2}{gray!15}{white}
\begin{tabular*}{\linewidth}{@{\extracolsep{\fill}}rrcccrr@{}}
\toprule
$\boldsymbol{n}$ & \textbf{Shots} & \textbf{Mean $\Delta$} & \textbf{Std $\Delta$} & \textbf{CV} & \textbf{\#pos} & \textbf{\#neg} \\
\midrule
\multirow{5}{*}{1}
& 0  & $-1.8\%$ & $2.9\%$ & $1.6$ & 4  & 13 \\
& 1  & $+2.1\%$ & $1.1\%$ & $0.5$ & 16 & 1  \\
& 2  & $+3.0\%$ & $1.1\%$ & $0.4$ & 17 & 0  \\
& 10 & $+2.1\%$ & $0.5\%$ & $0.2$ & 17 & 0  \\
& 20 & $+1.4\%$ & $0.5\%$ & $0.3$ & 17 & 0  \\
\midrule
\multirow{4}{*}{30}
& 0  & $-2.3\%$ & $5.9\%$ & $2.5$ & 9  & 8 \\
& 2  & $+2.9\%$ & $1.0\%$ & $0.4$ & 17 & 0 \\
& 10 & $+1.2\%$ & $0.8\%$ & $0.7$ & 15 & 1 \\
& 20 & $+0.4\%$ & $0.8\%$ & $2.4$ & 12 & 5 \\
\midrule
\multirow{4}{*}{100}
& 0  & $+0.9\%$ & $5.2\%$ & $5.5$ & 12 & 5 \\
& 2  & $+1.6\%$ & $1.3\%$ & $0.8$ & 16 & 1 \\
& 10 & $+0.8\%$ & $1.2\%$ & $1.4$ & 12 & 5 \\
& 20 & $+0.1\%$ & $1.0\%$ & $7.4$ & 9  & 8 \\
\bottomrule
\end{tabular*}
\end{table}

\subsection{Task Details}
\label{app:task_details}

\paragraph{Multi-Digit Multiplication.} Most models solve $>90\%$ of 3-digit multiplication problems without reasoning, while 5-digit multiplication exceeds the capability of frontier models without CoT. We construct a 4-digit multiplication dataset of 1,000 randomly generated problems with a fixed seed of 12345. Each problem is of the form \texttt{What is XXXX times YYYY?}

\paragraph{Multi-Step Arithmetic.} Following \citet{greenblatt2024counting}, each problem consists of 5 to 7 nested arithmetic operations (addition, subtraction, multiplication, integer division, and modulus) with operands sampled uniformly from $[-99, 99]$, e.g., \texttt{((1 + 3) - (15 \% 7)) * 2}. We use a fixed random seed and evaluate all models on the same 1,000 generated problems.

\paragraph{Variable Counting in Code.} We provide the model with a short code snippet and ask it to count the number of distinct variables assigned a value. We evaluate all models on the same 1,000 generated problems with a fixed random seed.

\subsection{Prompting Details and Ablations}
\label{app:prompts}

For the cross-model evaluation in Table~\ref{tab:cross_model}, we use the following system prompt for all models:
\begin{verbatim}
You will be given a math problem. Answer immediately using the format
`Answer: [ANSWER]' where [ANSWER] is just the numerical answer, nothing
else. No explanation, no words, no reasoning, just the number.
\end{verbatim}
For Qwen3-235B evaluations, we use a minimal system prompt with the \texttt{/no\_think} prefix to suppress chain-of-thought reasoning. Filler tokens are prefilled in the assistant context after the question, followed by \texttt{Answer:}, and the model generates only the numerical answer.

Table~\ref{tab:prompting_results} reports key prompting ablations for Opus~4.5 on 4-digit multiplication. Filler tokens must appear \emph{after} the question: placing them before the question yields no benefit. In the generated pause configuration, the model produces the pause tokens itself rather than receiving them via prefill; generation is the most effective configuration, achieving $+10.2\%$ over baseline. Removing the system prompt hint for filler tokens has no effect on accuracy, confirming the uplift is not due to the model following instructions to use filler tokens.

\begin{table}[h]
\centering
\caption{\textbf{Prompting ablations for Opus~4.5} on 4-digit multiplication and multi-step arithmetic. $\Delta$ is the absolute change from the 10-shot no-filler baseline.}
\label{tab:prompting_results}
\small
\rowcolors{2}{gray!15}{white}
\begin{tabular}{@{}lcr@{}}
\toprule
\textbf{Configuration} & \textbf{Accuracy} & $\boldsymbol{\Delta}$ \\
\midrule
\multicolumn{3}{l}{\textbf{Multiplication}} \\
\midrule
10-shot baseline          & $71.8_{\pm 1.4}$ & ---     \\
Prefilled pause (after)   & $80.7_{\pm 1.2}$ & $+8.9\%$ \\
Prefilled pause (no hint) & $80.8_{\pm 1.2}$ & $+9.0\%$ \\
Prefilled pause (before)  & $73.4_{\pm 1.4}$ & $+1.6\%$ \\
Generated pause           & $82.0_{\pm 1.2}$ & $+10.2\%$ \\
\midrule
\multicolumn{3}{l}{\textbf{Arithmetic}} \\
\midrule
10-shot baseline          & $45.8_{\pm 1.6}$ & ---     \\
Prefilled pause (after)   & $52.5_{\pm 1.6}$ & $+6.7\%$ \\
Prefilled pause (no hint) & $51.7_{\pm 1.6}$ & $+5.9\%$ \\
Prefilled pause (before)  & $44.2_{\pm 1.6}$ & $-1.6\%$ \\
Generated pause           & $53.5_{\pm 1.6}$ & $+7.7\%$ \\
\bottomrule
\end{tabular}
\end{table}

\subsection{Token Calibration and Selection Effects}
\label{app:calibration}

\paragraph{Token calibration.} Different filler types have different word-to-token ratios: a single word may map to one token (e.g., ``pause'') or multiple tokens (e.g., large Fibonacci numbers). We calibrate the number of filler words per type so that the resulting sequence contains approximately $n$ tokens under the target model's tokenizer. For each filler type and model, we perform a binary search over word count, counting tokens using the model's tokenizer (Anthropic token-counting API for Claude models, local tokenizer weights for open-weight models). For the cross-model comparison, counting tokens (1--100) tokenize consistently to approximately 200 tokens across all models evaluated.

\paragraph{Selection effects.} Some models disobey the system prompt and produce a chain-of-thought before answering, particularly on harder problems. We discard these samples, which can inflate accuracy by selecting for easier problems. When prompted to generate counting filler tokens, Opus~4.5 and Opus~4.6 produce explicit reasoning traces in roughly $60\%$ and $90\%$ of outputs, respectively. We address this by prefilling the answer prefix in the assistant context, which prevents the model from generating a CoT. This approach fully controls selection effects for Opus~4.5 but cannot be applied to Opus~4.6, GPT-5.2, or GPT-5.5, as their APIs do not permit assistant prefilling. Results for these models in Table~\ref{tab:cross_model} are marked with an asterisk.

\begin{table}[h]
\centering
\caption{\textbf{Mitigating selection effects on arithmetic.} Reasoning rate and accuracy under different configurations for Opus~4.5 and Opus~4.6.}
\label{tab:selectioneffects}
\small
\rowcolors{2}{gray!15}{white}
\begin{tabular}{@{}lcc@{}}
\toprule
\textbf{Configuration} & \textbf{Reasoning \%} & \textbf{Accuracy} \\
\midrule
\multicolumn{3}{l}{\textbf{Opus 4.5}} \\
\midrule
Generating 1--100           & $61.1$ & $67.4_{\pm 2.4}$ \\
$2\times$ few-shot examples & $0.2$  & $58.3_{\pm 1.6}$ \\
Assistant prefill (1--100)  & $0.3$  & $57.0_{\pm 1.6}$ \\
\midrule
\multicolumn{3}{l}{\textbf{Opus 4.6}} \\
\midrule
Generating 1--100           & $90.4$ & $90.7_{\pm 3.1}$ \\
$2\times$ few-shot examples & $68.2$ & $80.8_{\pm 2.2}$ \\
Assistant prefill (1--100)  & ---    & ---              \\
\bottomrule
\end{tabular}
\end{table}

\section{Filler Token Sweep}
\label{app:filler_sweep}

\subsection{Filler Token Examples}
\label{app:filler_examples}

Table~\ref{tab:filler_examples} lists the 17 filler token types used in our evaluations, with an example of each sequence.

\begin{table}[ht]
\centering
\caption{\textbf{Filler token types and examples.} Each sequence is calibrated to contain approximately $n$ tokens under the target model's tokenizer.}
\label{tab:filler_examples}
\small
\rowcolors{2}{gray!15}{white}
\begin{tabular}{@{}ll@{}}
\toprule
\textbf{Type} & \textbf{Example (truncated)} \\
\midrule
Counting       & 1 2 3 4 5 6 \ldots \\
Fibonacci      & 1 1 2 3 5 8 13 21 \ldots \\
Prime numbers  & 2 3 5 7 11 13 17 \ldots \\
Square numbers & 1 4 9 16 25 36 49 \ldots \\
Pi digits      & 3 1 4 1 5 9 2 6 5 \ldots \\
Random numbers & 47 83 12 91 55 38 \ldots \\
Digit words    & one two three four five \ldots \\
Alphabet       & a b c d e f g h \ldots \\
NATO phonetic  & alpha bravo charlie delta \ldots \\
Animals        & cat dog bear wolf lion \ldots \\
Fruits         & apple mango peach grape \ldots \\
Colors         & red blue green yellow \ldots \\
US states      & Alabama Alaska Arizona \ldots \\
Lorem ipsum    & lorem ipsum dolor sit amet \ldots \\
Pause          & pause pause pause pause \ldots \\
Ellipses       & \ldots\ \ldots\ \ldots\ \ldots\ \ldots \\
Random tokens  & xQ7 bR2 mK9 wP4 \ldots \\
\bottomrule
\end{tabular}
\end{table}

\subsection{Qwen3.5-397B}
\label{app:qwen35_sweep}

Qwen3.5-397B achieves $0.0\%$ at zero-shot and $90.6\%$ at 10-shot on 4-digit multiplication. We serve the model locally for this sweep, whereas Table~\ref{tab:cross_model} queries the model through OpenRouter; the OpenRouter deployment reaches a $99.3\%$ baseline on the same problems, so serving configuration substantially affects baseline accuracy. Table~\ref{tab:qwen35_sweep} reports the full filler token sweep at $n{=}100$.

\paragraph{Zero-shot.} The model scores $0.0\%$ on all conditions, including with filler tokens. Filler tokens provide no benefit when the model cannot solve the task without them.

\paragraph{Ten-shot at $n{=}100$.} Most filler types are catastrophically harmful. Word-based types (animals, nato, fruits, colors) drop accuracy from $90.6\%$ to near $0\%$. Only three types show marginal positive effects: Fibonacci ($+1.5\%$), ellipsis ($+1.4\%$), and pi digits ($+1.2\%$).

\paragraph{Mixed filler.} We tested whether catastrophic failure is a threshold or proportional effect by mixing Fibonacci tokens with animal tokens at varying ratios. Any mixture with $\geq 25\%$ animal tokens caused the model to produce extremely long outputs that timed out the server, even at $N{=}1{,}000$. The disruption is not just wrong answers: animal tokens cause the model to violate the concise answer format established by few-shot examples.

\paragraph{Arithmetic.} Qwen3.5-397B achieves $0.9\%$ at 10-shot on multi-step arithmetic. Filler tokens have no effect ($\Delta \approx 0$ across all types). The same model that shows catastrophic type sensitivity on multiplication shows no filler response on arithmetic, confirming that the filler token effect requires a minimum baseline capability.

\begin{table}[t]
\centering
\caption{\textbf{Full filler token sweep for Qwen3.5-397B on 4-digit multiplication} ($n{=}100$, $N{=}1{,}000$). Baseline: $0.0\%$ (0-shot), $90.6\%$ (10-shot). Most filler types are catastrophically harmful at 10-shot.}
\label{tab:qwen35_sweep}
\small
\rowcolors{2}{gray!15}{white}
\begin{tabular}{@{}lcc@{}}
\toprule
\textbf{Type} & \textbf{0-shot $\Delta$} & \textbf{10-shot $\Delta$} \\
\midrule
fibonacci       & $+0.0$ & $+1.5$  \\
ellipsis        & $+0.0$ & $+1.4$  \\
pi\_digits      & $+0.0$ & $+1.2$  \\
counting        & $+0.0$ & $-1.4$  \\
squares         & $+0.0$ & $-9.8$  \\
primes          & $+0.0$ & $-10.0$ \\
alphabet        & $+0.0$ & $-16.3$ \\
random\_numbers & $+0.0$ & $-16.0$ \\
pause           & $+0.0$ & $-78.9$ \\
random\_tokens  & $+0.0$ & $-89.1$ \\
states          & $+0.0$ & $-89.7$ \\
digit\_words    & $+0.0$ & $-89.8$ \\
lorem           & $+0.0$ & $-89.9$ \\
colors          & $+0.0$ & $-90.4$ \\
fruits          & $+0.0$ & $-90.6$ \\
nato            & $+0.0$ & $-90.6$ \\
animals         & $+0.0$ & $-90.6$ \\
\bottomrule
\end{tabular}
\end{table}

\section{Mechanistic Analysis}
\label{app:mechinterp}

We run three mechanistic experiments on Qwen3-235B on the variable counting task, comparing animals filler tokens (which provide strong uplift) against Fibonacci tokens (which provide weak uplift). All experiments use 100 filler tokens in the 10-shot setting.

\paragraph{Filler token order.} We fix the total token count at 100 and vary the ordering of mixed sequences. The ``animals$\to$Fibonacci'' ordering (50 animals followed by 50 Fibonacci) recovers most of the accuracy from pure animals tokens, while ``Fibonacci$\to$animals'' performs worse than pure Fibonacci. Early positions in the filler sequence are disproportionately important.

\paragraph{Activation patching.} We patch residual stream activations from animals filler tokens into a forward pass that otherwise uses Fibonacci tokens, measuring recovery of the performance gap: $(\text{patched} - \text{Fibonacci}) / (\text{animals} - \text{Fibonacci})$. We apply the patch across all 94 layers at the full filler span. Patching a single token position yields no meaningful recovery, confirming that the useful computation is distributed across the full sequence.

\begin{table}[h]
\centering
\caption{\textbf{Activation patching recovery by layer} (Qwen3-235B, variable counting). Values $>1$ reflect mild overshoot, consistent with activation patching literature.}
\label{tab:act_patch}
\small
\rowcolors{2}{gray!15}{white}
\begin{tabular}{@{}lcccccccccc@{}}
\toprule
\textbf{Layer} & 0 & 10 & 20 & 30 & 40 & 50 & 60 & 70 & 80 & 90 \\
\midrule
Recovery & 0.95 & 0.93 & 1.11 & 1.14 & 0.79 & 0.54 & 0.40 & $-0.13$ & $-0.07$ & $-0.06$ \\
\bottomrule
\end{tabular}
\end{table}

Patching at layers 0--30 recovers over 90\% of the performance gap; patching at layers 70 and above yields negative recovery. The benefit is established in early layers, and late-layer activations from animals tokens are harmful when inserted into a Fibonacci context.

\paragraph{Linear probing.} We train linear probes on mean-pooled residual stream activations across the filler span at each layer to predict the correct answer. Probing a single token position yields no meaningful signal; mean pooling across the full span is required, consistent with the activation patching result that computation is distributed across the sequence.

\begin{table}[h]
\centering
\caption{\textbf{Linear probe accuracy by layer} (Qwen3-235B, variable counting). Chance is $20\%$. Probes are trained on mean-pooled residual stream activations across the filler span.}
\label{tab:linear_probe}
\small
\rowcolors{2}{gray!15}{white}
\begin{tabular}{@{}lcccccccc@{}}
\toprule
\textbf{Layer} & 0 & 15 & 30 & 45 & 60 & 75 & 90 \\
\midrule
Baseline (no filler) & $20.0\%$ & $48.7\%$ & $53.3\%$ & $41.3\%$ & $56.7\%$ & $56.0\%$ & $48.0\%$ \\
Animals              & $20.0\%$ & $59.3\%$ & $67.3\%$ & $60.7\%$ & $65.3\%$ & $70.0\%$ & $69.3\%$ \\
Fibonacci            & $20.0\%$ & $50.0\%$ & $54.0\%$ & $51.3\%$ & $53.3\%$ & $54.7\%$ & $44.7\%$ \\
\bottomrule
\end{tabular}
\end{table}

Animals filler tokens encode substantially more task-relevant information than Fibonacci tokens at every layer from layer 15 onward. The animals probe reaches $70\%$ accuracy at layer 75, compared to $54.7\%$ for Fibonacci and $56.0\%$ for the no-filler baseline, indicating that the choice of filler token type determines how effectively task-relevant information accumulates in the residual stream.

\subsection{Attention Masking}
\label{app:attention_masking}

We run attention masking experiments on Qwen3-235B on 4-digit multiplication (10-shot, $n{=}100$, $N{=}1{,}000$) to probe which attention pathways carry the filler token benefit. Each mask blocks a designated query region from attending to a designated key-value region; all other attention is unmodified.

We test two primary masks: Mask~A blocks the final filler span from attending to the full few-shot block; Mask~B blocks it from all prior context. We also run source-region ablations that isolate specific key-value regions and a scoring-only mask that blocks gold answer tokens from attending to the final filler span.

\paragraph{Core masking results.} Table~\ref{tab:mask_core} shows that applying either mask causes only modest changes in accuracy ($\leq 1$--$2\%$) and log-probability gain. Blocking the entire few-shot block or all prior context does not remove the uplift.

\begin{table}[h]
\centering
\caption{\textbf{Core attention masking results} (Qwen3-235B, 4-digit multiplication). Mask~A blocks the filler span from the few-shot block; Mask~B blocks it from all prior context.}
\label{tab:mask_core}
\small
\rowcolors{2}{gray!15}{white}
\begin{tabular}{@{}lcc@{}}
\toprule
\textbf{Condition} & \textbf{Accuracy} & \textbf{$\Delta \log p$} \\
\midrule
Baseline (10-shot)        & $70.3\%$ & $0.000$ \\
pi\_digits unmasked       & $69.8\%$ & $+0.329$ \\
pi\_digits Mask~A         & $69.1\%$ & $+0.329$ \\
counting unmasked         & $67.9\%$ & $+0.085$ \\
counting Mask~A           & $67.9\%$ & $+0.086$ \\
\midrule
Baseline (0-shot)         & $57.3\%$ & $0.000$ \\
counting 0-shot unmasked  & $60.9\%$ & $+0.630$ \\
counting 0-shot Mask~B    & $60.7\%$ & $+0.641$ \\
\bottomrule
\end{tabular}
\end{table}

\paragraph{Source-region ablations.} Table~\ref{tab:mask_source} reports finer-grained results. For 0-shot counting, blocking filler self-attention reduces accuracy more than blocking the final question, suggesting tokens within the span build on each other. For 10-shot random\_numbers, blocking the few-shot block causes the largest drop, while blocking the final question has no effect. No single pathway accounts for the full benefit, and results differ between filler types.

\begin{table}[h]
\centering
\caption{\textbf{Source-region ablations} (Qwen3-235B, 4-digit multiplication). Each mask blocks the final filler span from one key-value region.}
\label{tab:mask_source}
\small
\rowcolors{2}{gray!15}{white}
\begin{tabular}{@{}lccc@{}}
\toprule
\textbf{Blocked region} & \textbf{Baseline} & \textbf{Unmasked} & \textbf{Masked} \\
\midrule
\multicolumn{4}{l}{\textit{0-shot counting}} \\
All prior context      & $41.8\%$ & $47.2\%$ & $48.8\%$ \\
Final question         & $42.1\%$ & $50.4\%$ & $49.5\%$ \\
Filler self-attention  & $42.0\%$ & $50.8\%$ & $48.1\%$ \\
System prompt          & $40.2\%$ & $48.7\%$ & $47.9\%$ \\
\midrule
\multicolumn{4}{l}{\textit{10-shot random\_numbers}} \\
Few-shot block         & $64.0\%$ & $67.6\%$ & $66.1\%$ \\
Final question         & $63.9\%$ & $67.0\%$ & $67.5\%$ \\
Filler self-attention  & $64.9\%$ & $65.7\%$ & $66.1\%$ \\
All prior context      & $64.3\%$ & $66.4\%$ & $68.4\%$ \\
System prompt          & $64.0\%$ & $66.1\%$ & $64.2\%$ \\
\bottomrule
\end{tabular}
\end{table}

The masking results are consistent with the activation patching and probing findings: the filler token benefit is distributed across the sequence, type-specific, and not reducible to a single dominant attention pathway. The zero-shot counting result in Table~\ref{tab:mask_core} further shows that uplift can persist when the filler span cannot attend to any prior context, implying that in some settings the benefit arises from mechanisms other than question-specific computation within the span.

\section{Hidden Goal Results for Qwen3-235B}
\label{app:hidden_motive_qwen}

Table~\ref{tab:qwen_hidden_full} reports the full breakdown of hidden goal performance for Qwen3-235B on 4-digit multiplication ($N{=}1{,}000$), showing both NA accuracy (the rate at which the model correctly outputs N/A on constraint-satisfying problems) and non-NA accuracy (primary task performance on the remaining problems).

The baseline model never outputs N/A, confirming there are no false positives without filler tokens. With self-generated counting filler tokens on the easy constraint ($x \bmod 2 = 1$), the model correctly outputs N/A on $98.8\%$ of constraint-satisfying problems. Non-NA accuracy drops from $69.4\%$ to $52.1\%$, indicating the model is not outputting N/A indiscriminately: it applies the hidden objective to the correct subset, but at a cost to primary task accuracy. For the medium constraint ($x \bmod 3 = 2$), NA accuracy reaches $39.0\%$ with a smaller drop in non-NA accuracy ($69.8\% \to 44.5\%$). The hard constraint ($4x \bmod 7 = 5$) shows no hidden goal behavior at all: the model never outputs N/A, suggesting that computing $4x \bmod 7$ exceeds its latent modular arithmetic capability. The model pursues hidden objectives only when the constraint is within its latent capability.

\begin{table}[h]
\centering
\caption{\textbf{Full hidden goal breakdown for Qwen3-235B} on 4-digit multiplication ($N{=}1{,}000$). NA accuracy measures constraint satisfaction on problems where the constraint holds; non-NA accuracy measures multiplication accuracy on the remaining problems.}
\label{tab:qwen_hidden_full}
\small
\begin{tabular}{@{}llccc@{}}
\toprule
\textbf{Constraint} & \textbf{Mode} & \textbf{Overall} & \textbf{NA Acc.} & \textbf{Non-NA Acc.} \\
\midrule
\multirow{2}{*}{$x \bmod 2 = 1$ (easy)}   & Baseline  & $52.5\%$ & $0.0\%$  & $69.4\%$ \\
                                            & Generated & $63.5\%$ & $98.8\%$ & $52.1\%$ \\
\midrule
\multirow{2}{*}{$x \bmod 3 = 2$ (medium)} & Baseline  & $54.6\%$ & $0.0\%$  & $69.8\%$ \\
                                            & Generated & $43.3\%$ & $39.0\%$ & $44.5\%$ \\
\midrule
\multirow{2}{*}{$4x \bmod 7 = 5$ (hard)}  & Baseline  & $60.3\%$ & $0.0\%$  & $68.4\%$ \\
                                            & Generated & $60.0\%$ & $0.0\%$  & $68.1\%$ \\
\bottomrule
\end{tabular}
\end{table}

\section{RL Training Details}
\label{app:training_details}

This section provides a formal description of the reinforcement learning objective used for the RL experiments in Section~\ref{sec:rl}.

\subsection{Task and Prompt Format}

For multiplication, Each training problem asks the model to compute the product of two 4-digit integers.
The prompt includes 10 few-shot examples, each demonstrating the expected output format: a sequence of filler tokens followed by \texttt{Answer:~<number>}.
The few-shot examples contain the same number of filler tokens as the model is expected to generate.
The model generates its response autoregressively, producing filler tokens and then an answer.

\subsection{Reward}

Each completion receives a binary exact-match reward:
\begin{equation}
\label{eq:reward}
R_j = \mathbbm{1}[\hat{y}_j = y]
\end{equation}
where $\hat{y}_j$ is the integer extracted from sample $j$'s output and $y$ is the ground-truth product.

\subsection{Advantage Estimation}
\label{sec:advantage}

We follow a GRPO-style~\citep{shao2024deepseekmath} group advantage scheme.
For each problem $i$ in a batch of $B$ problems, we sample $G$ completions and compute the group mean reward $\bar{R}_i = \frac{1}{G}\sum_{j=1}^{G} R_{ij}$. For the experiments in Section~\ref{sec:rl} $B=32, G=32$.

\paragraph{MaxRL normalization.}
Rather than normalizing by the standard deviation as in GRPO, we normalize by the mean reward~\citep{tajwar2026maximumlikelihoodreinforcementlearning}:
\begin{equation}
\label{eq:advantage}
\hat{A}_{ij} = \frac{R_{ij} - \bar{R}_i}{\bar{R}_i + \epsilon}
\end{equation}
with $\epsilon = 10^{-6}$.
This upweights problems where few samples succeed (low $\bar{R}_i$) and downweights problems where most samples already produce the correct answer.

\paragraph{Zero-advantage filtering.}
When all $G$ samples for a problem receive the same reward---either all correct or all incorrect---the advantages are identically zero and the problem contributes no gradient.
These problems are dropped from the training batch and are always replaced with fresh problems drawn from the remaining training set to ensure that the batch is always the same size.

\paragraph{Answer-only weighting.}
Advantages are applied only to tokens at and after the \texttt{Answer:} delimiter; filler tokens receive $\hat{A}_t = 0$ and produce no gradient signal.
The model therefore receives no direct optimization pressure on filler token content, yet RL still discovers useful filler strategies because the choice of filler affects answer-token probabilities across rollouts.

\subsection{Policy Gradient Objective}
\label{sec:policy_loss}

We optimize a PPO-style~\citep{schulman2017ppo} clipped policy gradient objective with asymmetric clipping and hard ratio masking.

\paragraph{Importance sampling ratio.}
Let $\pi_\theta$ denote the current policy and $\pi_{\mathrm{old}}$ the policy that generated the rollouts. The per-token importance sampling ratio is:
\begin{equation}
\label{eq:ratio}
r_t = \frac{\pi_\theta(a_t \mid s_t)}{\pi_{\mathrm{old}}(a_t \mid s_t)}
\end{equation}

\paragraph{Asymmetric PPO clipping.}
The per-token clipped surrogate loss is:
\begin{equation}
\label{eq:ppo_clip}
\mathcal{L}_t^{\mathrm{clip}} = -\min\!\bigl(r_t\,\hat{A}_t,\;\operatorname{clip}(r_t,\, 1 - \epsilon_{\mathrm{lo}},\, 1 + \epsilon_{\mathrm{hi}})\,\hat{A}_t\bigr)
\end{equation}
with $\epsilon_{\mathrm{lo}} = 0.2$ and $\epsilon_{\mathrm{hi}} = 0.28$.
The asymmetry permits the policy to increase probability on positively-advantaged tokens more freely (upper ratio bound $1.28$) than to decrease probability on negatively-advantaged tokens (lower bound $0.80$).

\paragraph{Hard ratio masking.}
Following \citet{glm5}, tokens where the importance sampling ratio deviates too far from unity receive zero gradient. Specifically, we define a per-token mask:
\begin{equation}
\label{eq:hard_mask}
m_t = \mathbbm{1}\!\Bigl[\frac{1}{\beta} \le r_t \le \beta\Bigr]
\end{equation}
with $\beta = 2.0$, so tokens with $r_t \notin [0.5, 2.0]$ are excluded from the gradient computation entirely. This provides a hard safety margin beyond the soft PPO clip.

\paragraph{Batch loss.}
The final loss averages over the set $\mathcal{V}$ of valid token positions (answer tokens with nonzero advantage, excluding padding):
\begin{equation}
\label{eq:loss}
\mathcal{L} = \frac{1}{|\mathcal{V}|}\sum_{t \in \mathcal{V}} m_t \cdot \mathcal{L}_t^{\mathrm{clip}}
\end{equation}

\paragraph{KL monitoring.}
We monitor the KL divergence between $\pi_\theta$ and $\pi_{\mathrm{old}}$ using the K3 estimator~\citep{schulman2017kl}:
\begin{equation}
\label{eq:k3}
\widehat{D}_{\mathrm{KL}}^{\mathrm{K3}} = \mathbb{E}_t\bigl[r_t - \log r_t - 1\bigr]
\end{equation}
This quantity is logged for diagnostics but is \emph{not} used as a penalty term in the loss.

\begin{figure}
    \centering
    \includegraphics[width=1\linewidth]{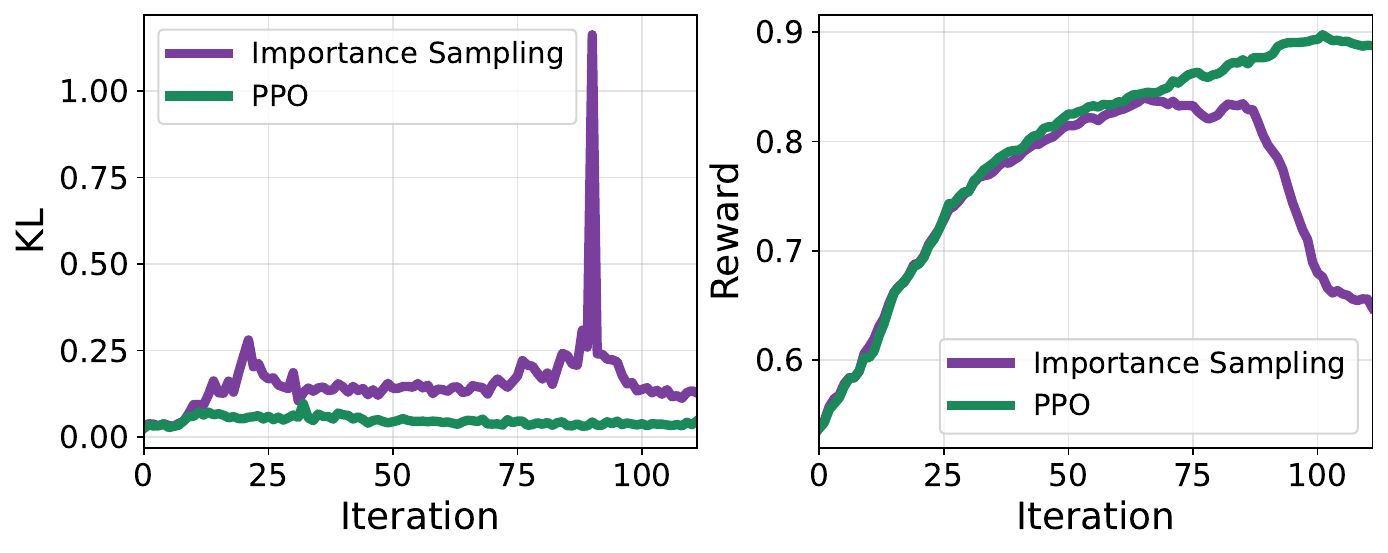}
    \caption{We compare the estimated KL and the mean reward for the PPO-style loss that we use, vs an importance-sampling-style loss function. These curves are for the model that chooses what kinds of filler tokens to generate, trained on multiplication~(Figure~\ref{fig:strategy-proportions}, left).}
    \label{fig:ppo-vs-is}
\end{figure}

In Figure~\ref{fig:ppo-vs-is} we compare our objective to an importance sampling objective and validate that the clipping and hard ratio masking serve to stabilize training so that the model continues improving past 75 iterations.

\subsection{Infrastructure}

\paragraph{Asynchronous RL.}
We use asynchronous RL~\citep{piche2025pipelinerlfasteronpolicyreinforcement,glm5} to overlap generation and training: while the current batch is being trained, rollouts for the next batch are generated concurrently on the inference server.
After each optimizer step, training weights are synchronized to the inference server, during which the inference server is briefly paused. We do not flush the KV cache after weight sync.

\paragraph{Rollout router replay.}
Because Qwen3-235B is a Mixture-of-Experts model, expert routing decisions are non-deterministic and differ between inference and training if not controlled.
We use rollout router replay~\citep{ma2025stabilizingmoereinforcementlearning}: expert routing indices from the inference forward pass are captured and replayed during the training forward pass, ensuring that the training gradient corresponds to the same expert assignments that produced the rollout.

\paragraph{Sampling.}
We use a modified version of SGLang for all rollouts.

\subsection{Hyperparameters}

\begin{table}[h]
\centering
\small
\begin{tabular}{@{}ll@{}}
\toprule
\textbf{Parameter} & \textbf{Value} \\
\midrule
Model & Qwen3-235B-A22B-Instruct-2507 \\
Training mode & Full-weight (all parameters) \\
Optimizer & Adam \\
Learning rate & $5 \times 10^{-6}$ \\
Gradient clip norm & 1.0 \\
Batch size $B$ & 32 problems \\
Group size $G$ & 32 completions per problem \\
PPO clip $\epsilon_{\mathrm{lo}}$ / $\epsilon_{\mathrm{hi}}$ & 0.2 / 0.28 \\
Hard mask $\beta$ & 2.0 \\
Few-shot examples & 10 \\
Advantage weighting & Answer-only \\
Advantage normalization & MaxRL \\
\bottomrule
\end{tabular}
\caption{Hyperparameters for the 4-digit multiplication RL experiments.}
\label{tab:rl_hyperparams}
\end{table}

\subsection{Prompts}

Model that learns to generate its own filler token sequences: (5.1)

\begin{verbatim}
Do not reason explicitly or show your work. 
Instead, you may generate up to {filler_tokens} filler tokens to help you 
process the problem internally. 
You may reason in an encoded manner through your choice 
and ordering of these filler tokens. 
These can be any of the following sequences:
{filler token sequences and descriptions here}
You may only generate up to {filler_tokens} tokens. 
Note that each digit you generate is a separate token.\n
Format:\n[filler tokens]\nAnswer: [number]
\end{verbatim}

Model that can only use random numbers filler: (5.2)

\begin{verbatim}
Do not reason explicitly or show your work. 
Instead, you may generate up to {filler_tokens} filler tokens to help you 
process the problem internally. 
You may reason in an encoded manner through your choice 
and ordering of these filler tokens.
These should be random numbers (e.g. "482 17 935 6 204 73...").\n
You may only generate up to {filler_tokens} tokens. 
Note that each digit you generate is a separate token.\n'
Format:\n[filler tokens]\nAnswer: [number]
\end{verbatim}

\section{RL Experiment Results}
\label{app:rl_results}

\subsection{Filler Token Preferences After RL}

We train Qwen3-235B with RL for 114 steps on 4-digit multiplication, allowing the model to generate filler tokens freely. Zero-shot baseline accuracy improves from $42.0\%$ to $66.5\%$ during training. RL alters the filler type hierarchy: counting ($+2.0\%$) and ellipsis ($+1.7\%$) become the most helpful types after training, while lorem ipsum ($-1.6\%$) and NATO ($-1.7\%$) become harmful despite being helpful in the base model ($+3.5\%$ and $+2.9\%$ respectively).

\paragraph{Test-time filler benefit.} After RL training, the model does not benefit from filler tokens at test time. At $N{=}10{,}000$: prefilled random numbers yield $73.62\%$ vs.\ baseline $73.19\%$ ($+0.43\%$, $+0.7\sigma$, not significant); model-generated random numbers yield $73.30\%$ ($+0.11\%$, $+0.2\sigma$, not significant). The RL training improves baseline accuracy but does not install a durable invisible reasoning capability that transfers to test time.

\begin{figure}[h]
\centering
\includegraphics[width=0.8\linewidth]{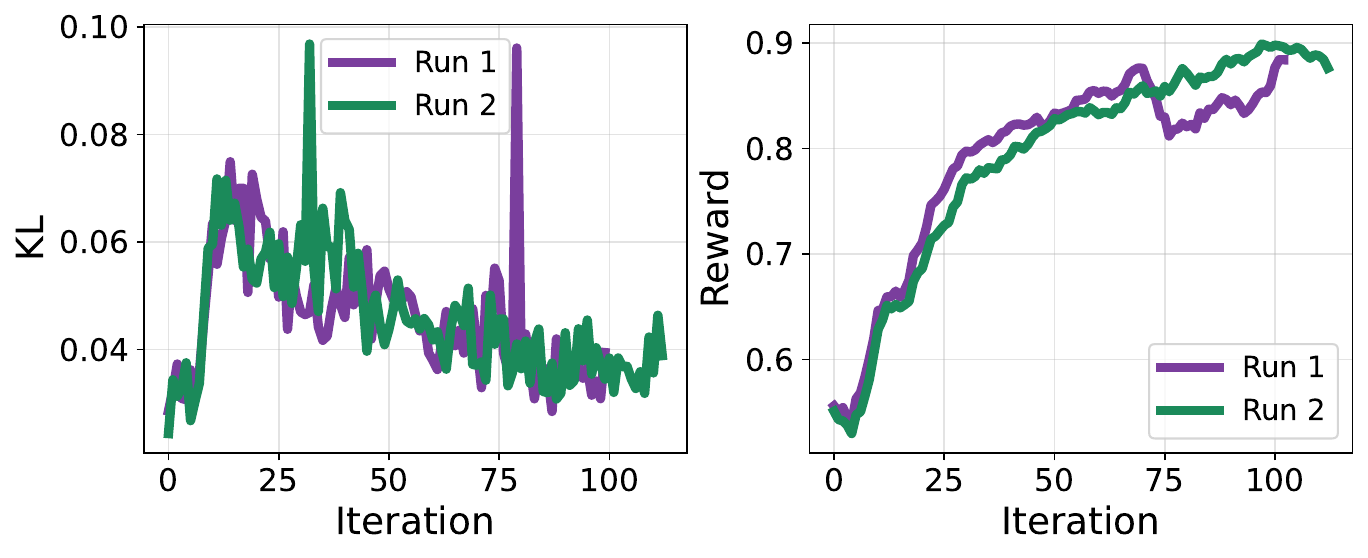}
\caption{\textbf{KL and reward curves for RL training on 4-digit multiplication.} The PPO-style clipped objective stabilizes training past 75 iterations, where an importance-sampling objective diverges.}
\label{fig:kl-reward-mult}
\end{figure}

\section{Supervised Fine-Tuning Details}
\label{app:sft}

We construct SFT datasets for Qwen3-235B by generating filler token sequences from Opus~4.5. Experiments vary along three axes: filler token type (counting 1--100 or ellipses), supervision strategy (full output or answer-only), and training data (all synthetic arithmetic problems or the subset Opus~4.5 solves with filler tokens). Training uses learning rate $10^{-4}$ with linear decay, LoRA rank 32, and batch size 64; varying these parameters does not affect the results. The loss is standard cross-entropy with a supervision mask $w_i \in \{0, 1\}$:
\begin{equation}
\mathcal{L}_{\text{SFT}} = -\frac{1}{\sum_i w_i}\sum_{i=1}^{N} w_i \cdot \log p_\theta(y_i \mid y_{<i})
\end{equation}
where $w_i = 1$ for all tokens under full supervision, or only for tokens after \texttt{Answer:} under answer-only supervision.

Across all configurations, SFT fails to transfer invisible reasoning. After fine-tuning, baseline accuracy improves, but generating filler tokens provides no additional boost. The model learns to reproduce filler token sequences from training data, but any accuracy gain is also present without filler tokens. This is expected: the computation that makes filler tokens useful for Opus~4.5 occurs in latent space. The tokens themselves carry no transferable signal, so imitating them provides no benefit.

\end{document}